# DP-MDM: Detail-Preserving MR Reconstruction via Multiple Diffusion Models

Mengxiao Geng, Jiahao Zhu, Xiaolin Zhu, Qiqing Liu, Dong Liang, *Senior Member, IEEE*, Qiegen Liu, *Senior Member, IEEE*

*Abstract*—Detail features of magnetic resonance images play a crucial role in accurate medical diagnosis and treatment, as they capture subtle changes that pose challenges for doctors when performing precise judgments. However, the widely utilized naive diffusion model has limitations, as it fails to accurately capture more intricate details. To enhance the quality of MRI reconstruction, we propose a comprehensive detail-preserving reconstruction method using multiple diffusion models to extract structure and detail features in k-space domain instead of image domain. Moreover, virtual binary modal masks are utilized to refine the range of values in k-space data through highly adaptive center windows, which allows the model to focus its attention more efficiently. Last but not least, an inverted pyramid structure is employed, where the top-down image information gradually decreases, enabling a cascade representation. The framework effectively represents multi-scale sampled data, taking into account the sparsity of the inverted pyramid architecture, and utilizes cascade training data distribution to represent multi-scale data. Through a step-by-step refinement approach, the method refines the approximation of details. Finally, the proposed method was evaluated by conducting experiments on clinical and public datasets. The results demonstrate that the proposed method outperforms other methods.

## I. Introduction

Magnetic Resonance Imaging (MRI) has undeniably sparked a medical revolutiime is often lengthy and requires accelon. MRI provides high-resolution images with rich contrast [1], but its scanning teration. Two commonly used acceleration strategies are Parallel Imaging (PI) [2] and Compressed Sensing (CS) [3]. Generally, PI achieves acceleration factors of 2 or 3 without significant loss in image quality, while CS enables larger factors by under-sampling the k-space signals. However, these acceleration strategies may result in artifacts and decreased image quality, making the optimization of reconstruction processes for under-sampled data a focal point in MRI research.

With the rapid progress of deep learning techniques in image processing, new tools have emerged for MRI reconstruction. The main reconstruction methods can be broadly classified into two categories: model-driven and data-driven methods. Model-driven methods are founded on specific optimization equations, which are mathematically derived and then transformed into forms that are suitable for building deep neural networks. These methods integrate artificial neural networks (ANNs) [4] to form a subset of reconstruction techniques. On the other hand, data-driven methods focus on extracting patterns, relationships, and correlations from extensive datasets to establish models. Advancements in computer technology have propelled data-driven methods to significant breakthroughs in MRI reconstruction. For instance, the variational autoencoder (VAE) model [5] learns the latent representation of data through variational inference to achieve data reconstruction and generation. The generative adversarial networks (GAN) model [6]-[8] simulates data distribution through the adversarial learning process of a generator and a discriminator. The deep cascade of convolutional neural networks (DC-CNN) model [9]-[12] employs cascaded convolutional neural networks to progressively learn residuals for MRI reconstruction. The unrolled iterative reconstruction (UnrolledIR) model [13]-[15] combines iterative reconstruction methods with deep learning to reconstruct high-quality images from incomplete sampling data by unrolling the iterative process into a deep network. Furthermore, other models like PixelCNN [16], [17] and deep generative models [18] have been explored in MRI reconstruction, contributing to the development of the field.

In recent years, the diffusion model has emerged as a versatile approach that combines the strengths of both model-driven and data-driven methodologies. On one hand, the diffusion model is based on physical principles and assumptions to describe and explain the diffusion process. It utilizes a set of mathematical equations and parameters to capture the behavior and patterns of diffusion. On the other hand, the construction and parameter estimation of the diffusion model often rely on actual observed data, *i.e.*, the acquired k-space data. The model parameters are continuously optimized or calibrated through training to better fit the observed data. In summary, the diffusion model not only has a solid theoretical foundation and physical interpretation, but also provides enhanced accuracy and adaptability to data.

As the popularity and unique benefits of diffusion models continue to grow, an increasing number of researchers are applying them to MRI reconstruction. Notably, Jalal *et al.* [19] were the first to apply a diffusion generative model-based compressed sensing reconstruction framework to MRI, utilizing a diffusion model based on fractional matching for sampling. In addition, Song *et al.* [20] improved the diffusion model based on the physical principles of solving the forward model for inverse problems, enhancing the generalization ability of the model for medical image reconstruction. Subsequently, Ye *et al.* [21], [22] conducted extensive research in accelerating MR imaging using diffusion models, achieving significant results. Meanwhile, Liu *et al.* [23], [24] focused on the original k-space domain and high-

This work was supported in part by National Natural Science Foundation of China under 62122033 and Key Research and Development Program of Jiangxi Province under 20212BBE53001. (M. Geng and J. Zhu are co-first authors.) (Corresponding author: Q. G. Liu)

M. Geng, X. Zhu, Q. Q. Liu and Q. G. Liu are with School of Information Engineering, Nanchang University, Nanchang 330031, China. ({mxiaogeng, xlzhu, qqliu}@email.ncu.edu.cn, liuqiegen@ncu.edu.cn)

J. Zhu is with School of Mathematics and Computer Sciences, Nanchang University, Nanchang 330031, China. (zhujiahao@email.ncu.edu.cn)

D. Liang is with Research Center for Medical AI, Shenzhen Institutes of Advanced Technology, Chinese Academy of Sciences, Shenzhen 518055, China. (dong.liang@siat.ac.cn)

dimensional space domain of MRI. Despite the promising progress in MRI reconstruction, naive diffusion model-based methods often pay less attention to building effective multi-scale representations and exhibit limited capability to capture detailed information and strike a balance between structure and detail features.

In this work, we propose a comprehensive **D**etail **P**reserving reconstruction method using **M**ultiple **D**iffusion **M**odels (DP-MDM). The primary concept behind DP-MDM is to employ multiple diffusion models for multi-scale learning of high-frequency information, enhancing the preservation of image details. During the diffusion process, virtual binary modality masks are employed to filter out low-frequency information, and the adjustable mask center size can capture details within different ranges, preserving these details during the reconstruction stage. It is worth noting that multiple diffusion models can be optimized for different data features, and through combined learning, they can better adapt to different scenarios and data distribution. The main contributions of this work can be summarized as follows:
- We propose a comprehensive detail preserving reconstruction method using multiple diffusion models for simultaneously representing structure and detail features for MRI reconstruction tasks.
- We employ virtual binary masks that refines the value range of k-space data through highly adaptable central windows. This explicit sign for MRI reconstruction enables the model to focus its attention more effectively and maximize the reconstruction performance of the model.
- We construct an inverse pyramid architecture in which the reconstruction information is progressively streamlined from top to bottom. The architecture effectively captures the features of multi-scale data by cascade sparse representation of the data distribution.

The remainder of this paper is exhibited as follows. Section II briefly introduces some relevant works in this study. Section III contains the key idea of the proposed method. The experimental results are shown in Section IV. Discussion and conclusion are given in Section V and Section VI.

## II. RELATED WORK

### A. Problem Formulation

Multi-coil parallel MRI image reconstruction in k-space domain is traditionally an optimization problem as follows:
$$y_c = Ax_c + \eta_c, c = 1, 2, \cdots, C. \quad (1)$$
Here $x_c \in \mathbb{R}^n$ is the medical k-space data of $c$-th coil to be reconstructed, $C$ is the total number of coils, $y_c \in \mathbb{R}^n$ represents the under-sampled k-space data, $A \in \mathbb{R}^{n \times n}$ is a diagonal matrix with diagonal elements of 1 or 0 depending on the sampling mode, and $\eta_c$ is the Gaussian noise.

In general, the sparse reconstruction model can accurately reconstruct images from highly under-sampled observations and address the issues of limited acquisition data and slow acquisition time in MRI. The primary objective of sparse reconstruction is to identify the sparse solution $x$ of the underdetermined linear equation (1), aiming to reconstruct the original images using highly under-sampled observations. The MRI reconstruction model can be expressed as an energy minimization problem:
$$\min_x \|Ax - y\|_2^2 + \lambda R(x), \quad (2)$$
where $\|\cdot\|_2^2$ represents the $l_2$ norm and $\|Ax - y\|_2^2$ term is the data fidelity term that simulates the relation of the given image to the unknown $x$. $R(x)$ is the regularization term that models the prior knowledge of $x$. $\lambda$ is a positive parameter used to balance these terms. Recently, various regularization methods are incorporated for MRI reconstruction, including $l_0$ regularization [25], $l_1$ regularization [26], $l_p$ regularization [27] and TV regularization [28], *etc*. Moreover, supervised end-to-end learning approaches often depend on a discriminative manner to acquire an implicit prior, resulting in restricted flexibility and robustness [29]. In this work, we turn to the explicit prior construction for MRI reconstruction via the score-based generative model.

### B. Score-based Generative Model with SDE

A score-based generative model is a popular method in generative modeling that leverages gradient scores to characterize the variation of probability density functions [39]. This model is combined with stochastic differential equations (SDE), which allows for accurate gradient estimation using neural networks and the SDE solver to create high-quality samples and capture complex structures.

More specifically, the complex data distribution can be transformed into a Gaussian noise distribution by gradually introducing noise into the diffusion process. The diffusion process $\{x(t)\}_{t=0}^T$ is indexed by a continuous-time variable $t \in [0, T]$. There is a dataset of *i.i.d.* samples (*i.e.*, $x(0) \sim p_0$) and a tractable form to generate samples efficiently (*i.e.*, $x(T) \sim p_T$), where $p_0$ and $p_T$ refer to the initial distribution of data and the data distribution after $T$ steps, respectively. Here, the diffusion process can be mathematically represented as the solution to the following SDE:
$$dx = f(x,t)dt + g(t)dw, \quad (3)$$
where $f \in \mathbb{R}^n$ and $g \in \mathbb{R}$ are the drift coefficient and the diffusion coefficient of $x(t)$, respectively. $w \in \mathbb{R}^n$ represents the standard Brownian motion.

By starting with samples of $x(T) \sim p_T$ and reversing the process, samples of $x(0) \sim p_0$ can be obtained. The reverse process can be expressed as the reverse-time SDE:
$$dx = \left[ f(x,t) - g^2(t) \nabla_x \log p_t(x) \right] dt + g(t) d\overline{w}, \quad (4)$$
where $dt$ is the infinitesimal negative time step, $\overline{w}$ is the reverse-time flow process of $w$, and $\nabla_x \log p_t(x)$ is the score of each marginal distribution. Since the true $\nabla_x \log p_t(x)$ of all $t$ is unknown, it can be estimated by training a time-dependent scoring network $s_\theta(x(t), t)$, *i.e.*, satisfying $s_\theta(x(t), t) \simeq \nabla_x \log p_t(x)$. Afterwards, samples can be generated using numerical solvers for stochastic differential equations, such as Euler-Maruyama and stochastic Runge-Kutta methods [30]. These general-purpose numerical methods enable efficient and accurate sample generation from the reverse-time SDE model.

## III. METHOD

### A. Motivation

Since the high-frequency information fused in MRI reconstruction contains complex details such as edges and textures, it is crucial for medical diagnosis and treatment.

The loss of high-frequency information during the MRI reconstruction stage can lead to images that are blurry or lack essential details, presenting challenges for analysis and assessment by medical professionals. In this work, a novel multi-diffusion model is proposed for comprehensive detail-preserving reconstruction in Fig. 1(a). Specifically, we utilize multiple diffusion models, namely $GM_1$ and $GM_k$ ( $k = 2,3,\cdots,K.$ ), to extract both structure and detail information from the k-space during the prior learning stage, thereby approximating the ideal images. Furthermore, the $GM_1$ model excels at learning the distribution of structural information, while $GM_k$ (where $k = 2,3,\cdots,K.$) employs virtual binary modality masks to selectively adjust their focus, honing in on the training of detailed features. Then, the aggregation of features from structure to detail in Fig. 1(b) during the reconstruction stage effectively preserves fine details and corrects the errors of single diffusion model.

Moreover, an inverted pyramid architecture is constructed, with the image information gradually diminishing from the top to the bottom. In Fig. 1(c), the proposed model is effectively guided by circular virtual binary masks to focus its attention on detail features. Notably, the design of these masks takes into consideration the circular distribution observed in the amplitude data of k-space. As a result, the proposed method closely mirrors the underlying data distribution, rendering it more cognitively sound. In addition, the architecture effectively captures the multi-scale features by using a cascade sparse representation of the data distribution. This strategy facilitates a progressively refined approximation of high-frequency details, thereby incrementally enhancing the representation of details from coarse to fine. Subsequently, detailed descriptions of the prior learning stage and the iterative reconstruction stage for the DP-MDM are provided in Sections III. B and III. C, respectively.

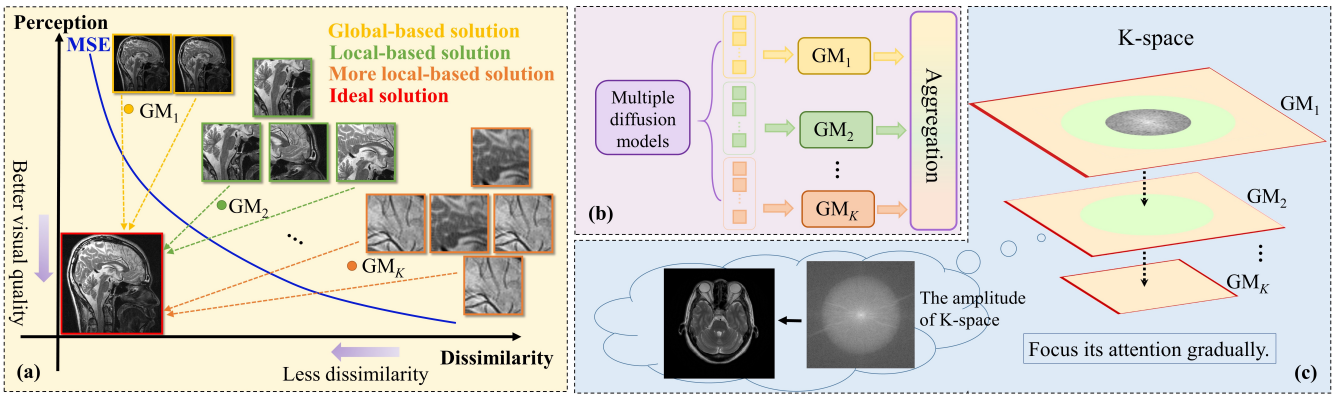

**Fig. 1.** The core idea of the proposed DP-MDM. $GM_1$ represents the data distribution for learning global structure information. $GM_k$ (where $k = 2,3,\cdots,K.$ ) represents data distribution that only learn local detail information. (a) Combining structure and detail features of an image to approximate the ideal solution. (b) DP-MDM uses multi-diffusion model embedding, allowing simulatneous representations of structure and detail features. (c) The inverted pyramid structure of DP-MDM. The data distribution information learned from $GM_1$ to $GM_k$ decreases sequentially.

*B. Prior Learning Stage*

During the prior learning stage, a novel multi-diffusion generalized reconstruction strategy is proposed to investigate the interrelation among k-space data. It is worth noting that the use of virtual binary modal masks, along with highly adaptive central windows, enables the model to localize the expression and processing of regional information within the k-space data. It enhances the ability of the model to effectively focus on retaining and capturing details. The construction method for these objects and their respective training models are shown in Fig. 2.

To begin with, a weighted matrix strategy is introduced for training model $GM_1$. This weighted strategy was initially proposed by Tu *et al.* [36]. By implementing this weighted strategy, it effectively constrains the size discrepancies in the k-space data and resolves the issue of significant variations in the value range. In other words, it is designed to deal with the imbalance issue of k-space data with both high-frequency and low-frequency information during the training stage. The weighted matrix operator can be expressed as follows:

$$x_1 = \hat{w} \odot I, \ \hat{w} = (r \cdot h^2 + r \cdot d^2)^p, \quad (5)$$

where $I$ denotes the initial data in k-space domain, and $\hat{w}$ is the specific weighted matrix. $\odot$ means element-wise multiplication and $r$ is introduced for setting the cut off value. $p$ decides the smoothness of the weight boundary while $h$ and $d$ are the count of frequency encoding lines and phase encoding lines.

Besides, multiple virtual mask strategies are designed specifically for training $GM_k$, where $k = 2,3,\cdots,K$. It is worth noting that, without losing generality, we chose $K = 3$ in the following work. For more discussion on the value of $k$, please refer to Section IV. C. These strategies work by employing a highly adaptable central window to restrict the value range of k-space data, thus facilitating the extraction of elusive high-frequency details while effectively filtering out unwanted low-frequency noise. Additionally, compared to $GM_2$, the $GM_3$ model has a lower risk of divergence and further enhances its ability to preserve details. Therefore, the $GM_3$ model plays a crucial role in improving reconstruction quality. These specific masked matrix operators are defined as follows:

$$x_2 = m_1 \odot I, \quad (6)$$

$$x_3 = m_2 \odot I, \quad (7)$$

where $m_1$ and $m_2$ are virtual circular masks of different diameters, with value 1 in k-space center (assume the window size $a$ is adjustable) and 0 for the rest of the region which denotes the high-frequency part. The specific mathematical description is as follows:

$$m_{\tilde{i}}(u,v) = \begin{cases} 1, & \text{if } d((u,v),(u_0,v_0)) < a/2 \\ 0, & \text{otherwise} \end{cases}, \ \tilde{i} = 1,2. \quad (8)$$

where $(u_0, v_0)$ denotes the center of the k-space data, $d(\cdot,\cdot)$ denotes a certain distance criterion. $a$ is the diameter of each circular virtual mask. Here, we use the Euclidean dis-tance, *i.e.*, a circle mask as default. The virtual masks differentiate the feature information across various regions within k-space data, which enables the model to focus its attention more effectively and mitigate the interference of redundant information on the output results, making more precise reconstruction possible. Note that the mask shape is not solely restricted to a circle one, more diverse forms of virtual binary modality masks will be given in Section III. D.

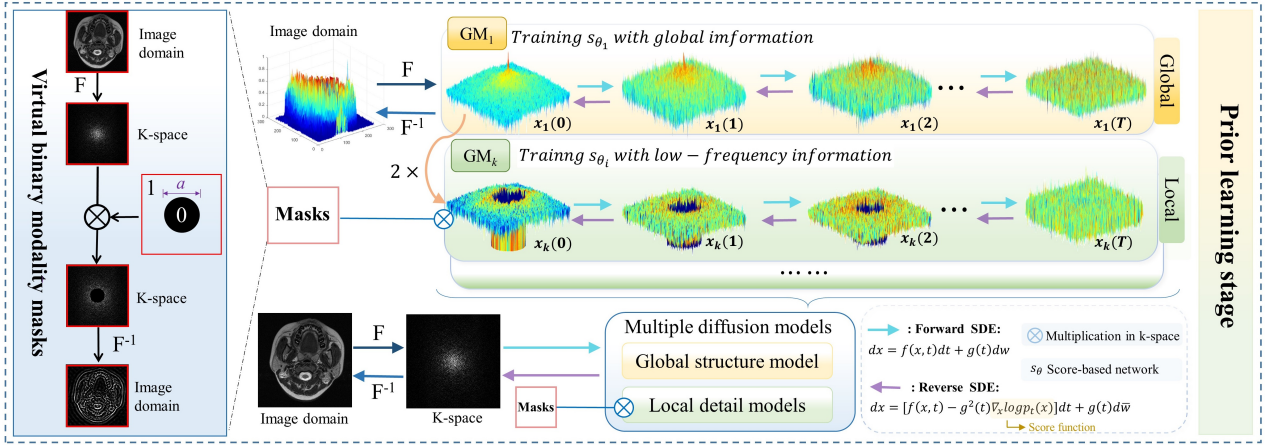

Fig. 2. The prior learning stage of DP-MDM. The stage is divided into multiple parts: $GM_1$, $GM_2$, $GM_3$. Specifically, $GM_1$ is used to learn the global distribution of structural information, while $GM_2$ and $GM_3$ are employed to learn the local distribution of detailed information.

In addition, the specific structures of $GM_1$, $GM_2$, and $GM_3$ all utilize score-based diffusion models, as detailed in Section II. These models consist of both forward SDE and reverse SDE processes. The forward process leverages SDEs to learn prior distribution by gradually injecting noise, facilitating a smooth transformation from complex data distribution to well-established prior distribution. Conversely, the reverse SDE converts noise back into data. The two processes are visually demonstrated in Fig. 2.

During the prior learning stage, the network can achieve its peak performance by optimizing the parameter $\theta^*$ of the score-based network. Assume that $x = x_1, x_2$ or $x_3$, the objective function can be described as follows:

$$\theta^* = \arg\min_\theta \mathbb{E}_t \{\lambda(t) \mathbb{E}_{x(0)} \mathbb{E}_{x(t)|x(0)}[ \\ \| s_\theta(x(t),t) - \nabla_x \log p_t(x(t)|x(0)) \|_2^2 ]\}, \quad (9)$$

where $\lambda(t)$ is the positive weighting function and $t$ is uniformly sampled over $[0,T]$. $p_t(x(t)|x(0))$ is the Gaussian perturbation kernel centered at $x(0)$. Once the network satisfies $s_\theta(x(t),t) \simeq \nabla_x \log p_t(x)$, $\nabla_x \log p_t(x)$ will be known for all $t$ by solving $s_\theta(x(t),t)$.

### C. Iterative Reconstruction Stage

During the iterative reconstruction stage, the model is guided to progressively integrate structural information and detailed features, thereby achieving a more accurate and comprehensive image reconstruction. To begin with, the k-space data $x$ is transformed from the MR image $\tilde{I}$ using the Fourier transform operation $F$. The corresponding projection relationship can be expressed as follows:

$$x = F(\tilde{I}). \quad (10)$$

In addition, $F^{-1}$ is the inverse Fourier transform operation.

Based on the principles of Section III. B on prior learning, an ensemble learning approach is employed in the machine learning framework to combine multiple models for enhanced performance. Specifically, the extraction of high-frequency information is utilized to form a multi-frequency prior. $GM_1$ is responsible for capturing global structure information, while $GM_2$ and $GM_3$ focus on learning local information. Additionally, the advantages of different operators are leveraged to enable their mutual integration, resulting in improved network topology interpretability and practical effectiveness. Fig. 3 illustrates that various combination methods are proposed, including (a) Cascade, (b) Hybrid 1, (c) Hybrid 2 and (d) Parallel. Through the ablation study in Section IV, the cascade method achieved superior results. Thus, the following study adopts the cascade combination during the reconstruction stage to conduct the structure-to-detail feature interaction.

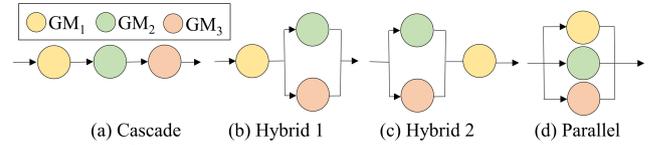

Fig. 3. Different combinations during the reconstruction stage.

Fig. 4 displays the overall framework of the proposed approach for MRI reconstruction. In the case of the cascade method, the under-sampled k-space data is obtained as input to the network by the Fourier transform. Then three inverse SDE solvers and data consistency modules are embedded for updating three collaborative works. These solvers leverage different prior information and undergo iterative updates to obtain full-sampled k-space data smoothly. Finally, the updated k-space data is reconstructed into an image by applying the inverse Fourier transform. It is worth noting that the reverse SDE solvers implemented here utilize the Predictor-Corrector sampler steps.

Furthermore, the overall framework during the iterative reconstruction stage can be described as follows:

$$\begin{cases} x_2^{i-1} = GM_2(x_1^i m_1), \\ x_1^{i-1} = x_1^i + m_1^H(x_1^i m_1 - x_2^{i-1}), \\ x_3^{i-1} = GM_3(x_1^{i-1} m_2), \\ x_1^{i-2} = x_1^{i-1} + m_2^H(x_1^{i-1} m_2 - x_3^{i-1}). \end{cases} \quad (11)$$

Here, $x_1$, $x_2$, and $x_3$ represent the output results of networks that utilize different prior information corresponding to $GM_1$, $GM_2$ and $GM_3$. $i$ denotes the total number of reverse iterations. The superscript $H$ is the Hermitian transpose operation. It is assumed that $x_1^i = GM_1(x_1^{i+1})$.

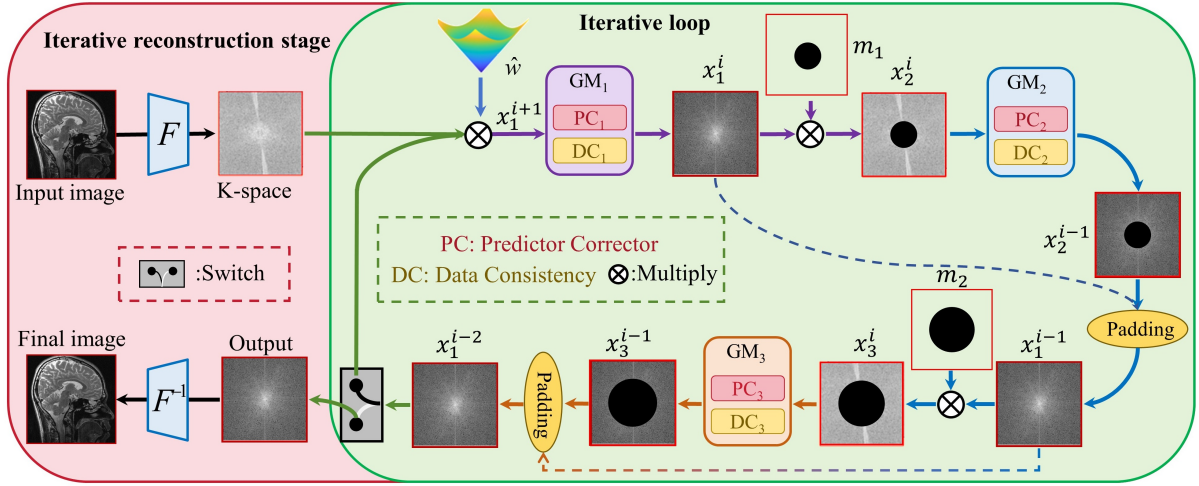

**Fig. 4.** The pipeline for iterative reconstruction stage of DP-MDM. The reconstruction stage adopts the cascade combination to conduct the structure-to-detail feature interaction.

Next, we will detail the Predictor-Corrector sampler and the data consistency steps during the reconstruction stage, respectively.

**Predictor-Corrector (PC):** Once the score network $s_\theta = s_{\theta_1}, s_{\theta_2}$ or $s_{\theta_3}$ is trained in Section III. B, it is inserted into Eq. (12) and solves the resulting reverse SDE to achieve MRI reconstruction:

$$dx = [f(x,t) - g^2(t)s_\theta(x,t)]dt + g(t)d\overline{w}, \quad (12)$$

where $s_{\theta_1}$, $s_{\theta_2}$ and $s_{\theta_3}$ respectively denote the training score networks associated with GM$_1$, GM$_2$, and GM$_3$. By the reverse SDE, random noise is transformed into data for sampling. To solve the reverse SDE, the PC sampling method is introduced at the sample updating step, as suggested in [31]. As part of the PC sampling, the predictor assumes the role of a numerical solver for the reverse SDE, specifically selected as the Variance Exploding (VE) SDE solver. Once the reverse-time SDE procedure is finalized, samples can be generated by utilizing the prior distribution. The predictor can be discretized as follows:

$$x^i = x^{i+1} + (\sigma_{i+1}^2 - \sigma_i^2)s_\theta(x_{i+1}, \sigma_{i+1}) + \sqrt{\sigma_{i+1}^2 - \sigma_i^2}z, \quad (13)$$

where $z \sim \mathcal{N}(0,1)$ refers to a standard normal distribution. $\sigma_i$ is the noise schedule at the $i$-th iteration.

As for the corrector, it refers to the annealed Langevin dynamics via transforming any initial sample $x(t)$ to the final sample $x(0)$ with the following iteration:

$$x^{i,j} = x^{i,j-1} + \varepsilon_i s_\theta(x^{i,j-1}, \sigma_i) + \sqrt{2\varepsilon_i}z, \quad (14)$$
$$j = 1, 2, \cdots, M, \quad i = N-1, \cdots, 0.$$

where $\varepsilon_i > 0$ is the step size and the above equation is repeated for $j = 1, 2, \cdots, M$, $i = N-1, \cdots, 0$. The theory of annealed Langevin dynamics guarantees that when $M \to \infty$ and $\varepsilon_i \to 0$, $x^i$ is sampled under designated conditions.

**Data Consistency (DC):** After each PC sampler step, a DC module is performed on the intermediate data. This module enforces data consistency during each iterative reconstruction step, aligning the output with the original k-space information. Hence, the minimization problem can be succinctly described as follows:

$$\min_x \{\|Ax - y\|_2^2 + \lambda \|x - x_{gen}\|_2^2\}, \quad (15)$$

where $x_{gen}$ is the entry in k-space generated by the network and $x$ is the medical k-space data to be reconstructed. Therefore, the corresponding solution $x^*$ can be solved through the first-order optimality condition:

$$x^*(u) = \begin{cases} x_{gen}(u), & \text{if } u \notin \Omega \\ [A^H y(u) + \lambda x_{gen}(u)]/(1+\lambda), & \text{if } u \in \Omega \end{cases} \quad (16)$$

where $\Omega$ denotes the specific locations in k-space that have been sampled. $x_{gen}(u)$ represents the entry at index $u$ in k-space domain generated by the network.

*D. Information Relationships of Virtual Binary Masks*

In this study, the virtual binary masks mentioned above play a critical role during the diffusion process. Different mask shapes focus on different information relationships. Next, further forms of virtual masks will be provided to explore a more general scenario.

Three virtual binary mask shapes are designed in Fig. 5. The aforementioned proposed method is only based on the DPMDM-I. The design of DPMDM-I includes constructing virtual binary masks $m_1$ and $m_2$, with their centers in a progressively layered circular form. The design of these masks takes into consideration the ability to separate high-frequency and low-frequency information, thereby enabling the model to concentrate on learning detail features. To further quantitatively analyze the relationship between the effective information, information entropy $E_1$ and $E_2$ are defined as follows:

$$E_1 = -\sum_{l=0}^{L-1} p_1(l) \cdot \log_2(p_1(l)), \quad (17)$$

$$E_2 = -\sum_{l=0}^{L-1} p_2(l) \cdot \log_2(p_2(l)), \quad (18)$$

where $L$ is the number of pixel values, $p_1(l)$ and $p_2(l)$ represent the probability of pixel value $l$ for date $x_2$ and $x_3$ of Eqs. (6) and (7), respectively. For DPMDM-I, these masks correspond to a contained relationship between the extracted information $E_1$ and $E_2$. Moreover, it is evident from Fig. 5 that the information entropy of $E_1+E_2$ in DPMDM-I is the highest, indicating that the model training process covers more information.

The remaining two methods DPMDM-II and DPMDM-III represent the intersected and disjoint relationships between the effective information $E_1$ and $E_2$. Specially, DPMDM-II employs a progressively layered radial form to construct virtual binary masks, aiming to eliminate both low-frequency and directional high-frequency information upon image introduction during the prior learning stage and reduce artifacts generated during the reconstruction stage. Conversely, DPMDM-III utilizes a random disjoint approach to create virtual binary masks, primarily retaining a minimal amount of high-frequency information, which can enhance hierarchical diversity among multiple diffusion models. In summary, by associating information entropy with these relationships, the assessment of data utilization can be more comprehensive during the training stage, thereby enhancing the accuracy of evaluating the learning effectiveness and generalization ability of the proposed model. Experiments in Section IV. B will further discuss this validity. The results demonstrate that circle and radial mask shapes perform better than the random mask shape.

| Method | $m_1$ | $m_2$ | Information relationship | Mathematical expression | Information entropy of $E_1+E_2$ |
|---|---|---|---|---|---|
| DPMDM-I (Contained) | | | | $E_2 \subset E_1$ | 3.5290 |
| DPMDM-II (Intersected) | | | | $E_1 \cap E_2 \neq \emptyset$ and $E_2 \not\subset E_1$ | 3.2418 |
| DPMDM-III (Disjoint) | | | | $E_1 \cap E_2 = \emptyset$ | 2.2333 |

**Fig. 5.** Construction of three different forms for virtual binary masks. $E_1$ and $E_2$ denote the amount of valid information contained in the k-space data $x_2$ and $x_3$, respectively. $\emptyset$ denotes the empty set.

### E. Overview of the Proposed DP-MDM

The efficient DP-MDM algorithm is proposed for MRI reconstruction, as seen in Algorithm 1. During the prior learning stage, three diffusion models are employed to obtain sufficient prior information by inputting different k-space data. The reconstruction stage of DP-MDM utilizes the prior information from these three different data distributions through a cascade operation, enabling a progressive transfer of information from structure to detail features and enhancing the preservation of high-frequency details. Moreover, the reconstruction stage not only leverages the trained network for prediction but also applies correction measures to reduce errors. The predictor and corrector work collaboratively to generate the final sample. Additionally, after each operation of the predictor or corrector, a DC module is enforced to ensure the image quality.

---

**Algorithm 1: DP-MDM**

**Prior Learning Stage**

1: **Input:** K-space dataset $I$
2: $x = x_1, x_2$ or $x_3$ for the three train processes:
   $x_1 = \hat{w} \odot I$; $x_2 = m_1 \odot I$; $x_3 = m_2 \odot I$;
3: **Output:** Learned $s_{\theta_1}(x_1,t)$, $s_{\theta_2}(x_2,t)$, $s_{\theta_3}(x_3,t)$.

**Iterative Reconstruction Stage**

1: Set: $s_{\theta_1}, s_{\theta_2}, s_{\theta_3}, N, M, \sigma, \varepsilon$
2: $\{x_1^N, x_2^N, x_3^N\} \sim \mathcal{N}(0, \sigma_{max}^2 \mathbf{I})$
3: **For** $i = N-1$ **to** $0$ **do**
4:    **For** $n = 1$ **to** $3$ **do**
5:       Update $x_n^i \leftarrow Predictor(x_n^{i+1}, \sigma_i, \sigma_{i+1}, s_{\theta_n})$
6:       Update $x_n^i$ via Eq. (16) **(DC)**
7:       **For** $j = 1$ **to** $M$ **do**
8:          Update $x_n^{i,j} \leftarrow Corrector(x_n^{i,j-1}, \sigma_i, \varepsilon_i, s_{\theta_n})$
9:          Update $x_n^{i,j}$ via Eq. (16) **(DC)**
10:      **End for**
11:      $x_{n+1}^i \leftarrow x_n^i$
12:    **End for**
13:    $x_n^i \leftarrow x_n^{i+1}$
14: **End for**
15: Return final image $\tilde{I} = F^{-1}(x)$

---

## IV. EXPERIMENTS

### A. Experimental Setup

In this section, the performance of DP-MDM is compared with state-of-the-art algorithms across various acceleration factors and sampling patterns, including traditional methods SAKE [32], P-LORAKS [33], deep learning-based method EBMRec [34], and naive diffusion-based methods HGGDP [35] and WKGM [36]. Both the comparative experiments and ablation experiments confirm the effective reconstruction ability of DP-MDM. To ensure comparability and fairness in the experiments, all methods are tested on the same datasets. Open-source code related to this work is available at: https://github.com/yqx7150/DP-MDM.

**Datasets:** Experiments were conducted on multiple datasets, including the *SIAT* dataset, the *T1-weighted brain* dataset, the *T1 GE brain* dataset, the *T2 transverse brain* dataset and the *Fast-MRI* dataset, to verify the effectiveness and robustness of the proposed DP-MDM model. Specifically, the *SIAT* dataset was provided by Shenzhen Institute of Advanced Technology, Chinese Academy of Sciences. The dataset comprises 500 2D MR images, each with 12-channel complex-valued data and a resolution of pixels. Coil compression was employed to convert these images into single-coil images for training optimization and an additional 4000 single-coil images were augmented to improve the prior learning stage. The MR images were obtained from a healthy volunteer using a T2-weighted turbo spin echo sequence on a 3.0 $T$ Siemens Trio Tim MRI scanner. The imaging parameters included a repetition time (TR)/echo time (TE) of $6100/90$ $ms$, a field of view (FOV) of $220 \times 220$ $mm^2$, and a pixel size of $0.9 \times 0.9$ $mm^2$.

To further assess the ability of DP-MDM to generalize, we evaluated its performance on in-vivo datasets comprising diverse sequences. Firstly, the *T1-weighted brain* dataset was acquired from a healthy volunteer using an 8-channel joint-only coil on a 1.5 $T$ MRI scanner, employing a T1-weighted, 3D spoiled gradient echo sequence. Secondly, the *T1-GE brain* dataset was obtained using a 3.0 $T$ GE MRI scanner, capturing MR images with 8-channel complex-valued data. The FOV was $220 \times 220$ $mm^2$, and TR/TE was $700/11$ $ms$. Also, 12-channel *T2 transverse brain* images with a size of $256 \times 256$ were acquired with a 3.0 $T$ Siemens (Munich, Germany), whose FOV was $220 \times 220$ $mm^2$ and TR/TE $5000/91$ $ms$. Furthermore, a *Fast-MRI* dataset [37] was publicly available and includes images acquired using a total of 20 coils.

*Parameter Setting:* The proposed algorithm is implemented in the PyTorch framework. The data distribution is perturbed with Gaussian noise by adding noise values ranging from $\sigma_{min} = 0.01$ to $\sigma_{max} = 378$. An Adam optimizer is chosen with $\beta_1 = 0.9$ and $\beta_2 = 0.999$ to optimize the network. Besides, we utilize $r = 0.075$, $N = 1000$ and $M = 1$ for the reconstruction stage as default, unless specified otherwise. For other parameters, we empirically tune the parameters in their suggested ranges to give their best performances. Noted that only two models, WKGM and DP-MDM, are all trained in k-space domain while the others are in image domain. The experiments are conducted on an Ubuntu system with an Intel Core i7 3.70 GHz processor and 2 NVIDIA GeForce RTX 2080 12 GB GPUs.

*Evaluation Metrics:* To quantitatively measure the error caused by the proposed method, the Peak Signal-to-Noise Ratio (PSNR) and Structural Similarity (SSIM) [36] are used to evaluate the quality of reconstruction images. PSNR measures the ratio between the maximum possible power of a signal and the power of noise corruption, while SSIM quantifies the similarity between the original image and the reconstructed images. A higher PSNR value indicates better quality of the reconstructed image, and SSIM values range from 0 to 1, with values closer to 1 indicating a greater similarity to a full-sampled image.

### B. Reconstruction Comparison

*Comparison of Different Under-sampling Patterns:* To verify the effectiveness of the proposed DP-MDM, a comparative experiment is conducted with the traditional methods, deep learning-based method, and naive diffusion-based methods. Table I compares the MRI reconstruction performance with different under-sampling patterns (*e.g.*, 2D Poisson, 2D Random, and Uniform sampling) and acceleration factors (*e.g.*, ×8, ×10, and ×15). The results indicate that the Uniform sampling modes pose greater challenges for reconstruction methods compared to the Poisson and Random modes at the same acceleration factor of ×15. Nevertheless, regardless of the under-sampled modes and acceleration factors, the DP-MDM method consistently achieves the highest PSNR and SSIM values in image domain. Specifically, in the Random under-sampled mode with an acceleration factor of ×15, the proposed method achieves improvements of 11.48 dB and 5.64 dB over traditional methods SAKE and P-LORAKS, respectively. Furthermore, in the Poisson sampling mode with an acceleration factor of ×10, the proposed method outperforms the naive diffusion model methods HGGDP and WKGM by 8.12 dB and 5.82 dB, respectively.

Further, a visual comparison of the reconstruction quality is provided in Fig. 6. In general, the DP-MDM produces clearer MR images compared with other reconstruction methods, and the proposed method further brings in more details. The reconstruction results for the *T1-GE Brain* dataset under the Random and Uniform sampling modes, as well as under the acceleration factor of ×15, is presented. For example, when examining the enlarged area in the green box of the second row, it is evident that SAKE, P-LORAKS, and EBMREc reveal numerous missing signals in their reconstructions, along with significant noise interference. These reconstructed results are notably inferior to DP-MDM. The error maps, which subtract the reconstructed target images from the full-sampled, are also consistent with the quantitative comparisons in Table I, as the errors decrease after the multiple diffusion models is integrated.

TABLE I
PSNR AND SSIM COMPARISON WITH STATE-OF-THE-ART METHODS UNDER DIFFERENT SAMPLING PATTERNS WITH VARYING ACCELERATION FACTORS.

| *T1-GE Brain* | SAKE | P-LORAKS | EBMRec | HGGDP | WKGM | DP-MDM |
|---|---|---|---|---|---|---|
| 2D Poisson *R*=10 | 37.57/0.9290 | 36.05/0.8723 | 29.60/0.7263 | 32.80/0.8986 | 35.10/0.9444 | **40.92/0.9456** |
| 2D Poisson *R*=15 | 34.99/0.8975 | 32.83/0.7996 | 26.90/0.6876 | 30.94/0.8685 | 34.15/0.9318 | **39.11/0.9320** |
| 2D Random *R*=8 | 28.40/0.8322 | 34.07/0.8226 | 25.76/0.6508 | 31.46/0.8541 | 32.70/0.9097 | **38.23/0.9232** |
| 2D Random *R*=15 | 23.66/0.7300 | 29.50/0.7058 | 23.61/0.5465 | 29.13/0.8648 | 29.68/0.8889 | **35.14/0.8891** |
| Uniform *R*=10 | 24.68/0.7930 | 32.28/0.8016 | 25.30/0.6888 | 29.51/0.8841 | 29.76/0.8694 | **36.18/0.9208** |
| Uniform *R*=15 | 23.07/0.7352 | 28.81/0.7430 | 26.37/0.7346 | 29.48/0.9020 | 28.70/0.8465 | **29.93/0.9027** |

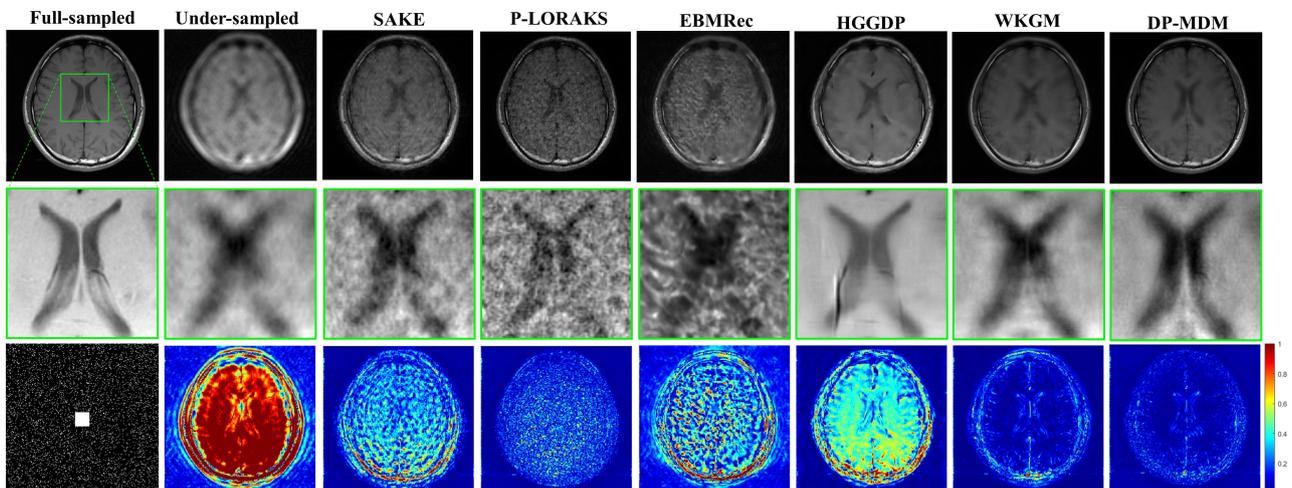

**Fig. 6.** Complex-valued reconstruction results at *R*=15 using 2D Random sampling with 8 coils. From left to right: Full-sampled, Under-sampled, reconstruction by SAKE, P-LORAKS, EBMRec, HGGDP, WKGM, and DP-MDM (OURS). The second row shows the enlarged view of the ROI (indicated by the green box in the first row), and the third row shows the error maps of the reconstruction.

*Different Datasets Under the Same Location:* In order to further prove the superiority of DP-MDM over other datasets, the results of different methods under different acceleration factors are shown in Table II. It can be seen that SAKE inevitably leads to the worst assessment results in most cases. Compared to SAKE, EBMRec brings an improvement, which means that automatically learning the intrinsic feature from the MR images is effective. In particular, P-LORAKS achieves better results than EBMRec in some cases, which indicates that the traditional P-LORAKS method is still competitive when the regularization term is chosen appropriately. Furthermore, the proposed DP-MDM gains the highest PSNR and SSIM scores. Compared to the second-best method, it provides a 1.95 dB higher PSNR and 2.08% higher SSIM under the 2D Poisson sampling with $R$=12. Overall, the DP-MDM proposed herein has the capacity to produce images with increased accuracy in pixel values and anatomical structures, thereby validating the enhancement of reconstruction quality achievable through the utilization of multiple diffusion methods.

TABLE II
PSNR AND SSIM COMPARISON WITH DIFFERENT METHODS UNDER POISSON AND RANDOM SAMPLING PATTERNS WITH VARYING ACCELERATION FACTORS.

| T1-weighted Brain | SAKE | P-LORAKS | EBMRec | HGGDP | WKGM | DP-MDM |
|---|---|---|---|---|---|---|
| 2D Random $R$=12 | 27.09/0.6460 | 30.29/0.7019 | 29.06/0.6689 | 29.85/0.7496 | 30.87/0.7170 | **37.14/0.8729** |
| 2D Random $R$=15 | 26.97/0.6360 | 28.89/0.6750 | 28.45/0.6701 | 29.40/0.7407 | 30.66/0.7159 | **36.25/0.8748** |
| **T2-transverse Brain** | **SAKE** | **P-LORAKS** | **EBMRec** | **HGGDP** | **WKGM** | **DP-MDM** |
| 2D Poisson $R$=8 | 31.75/0.8708 | 29.80/0.7823 | 25.84/0.8417 | 32.33/0.9257 | 30.06/0.8464 | **33.13/0.9310** |
| 2D Poisson $R$=12 | 22.37/0.8101 | 26.63/0.6853 | 25.33/0.7851 | 29.64/0.8855 | 28.31/0.8022 | **31.59/0.9063** |

To visually evaluate the performance of various reconstruction methods, results are shown on the Random and Poisson sampling modes with acceleration factor of ×12, respectively. In Fig. 7, the reconstruction results, regions of interest (ROI), and corresponding error maps from different algorithms are presented. P-LORAKS and EBMRec exhibit aliasing artifacts and blurred tissues, attributable to the loss of k-space data. Furthermore, naive diffusion-based methods HGGDP and WKGM obtain more promising results than EBMRec, consistent with the findings in Table II. However, these methods exhibit limitations in maintaining consistency between the reconstructed data and the under-sampled data in k-space, especially when compared to DP-MDM. As a result, DP-MDM produces images with notably enhanced detail and improved quality, as exemplified by the green box highlighting.

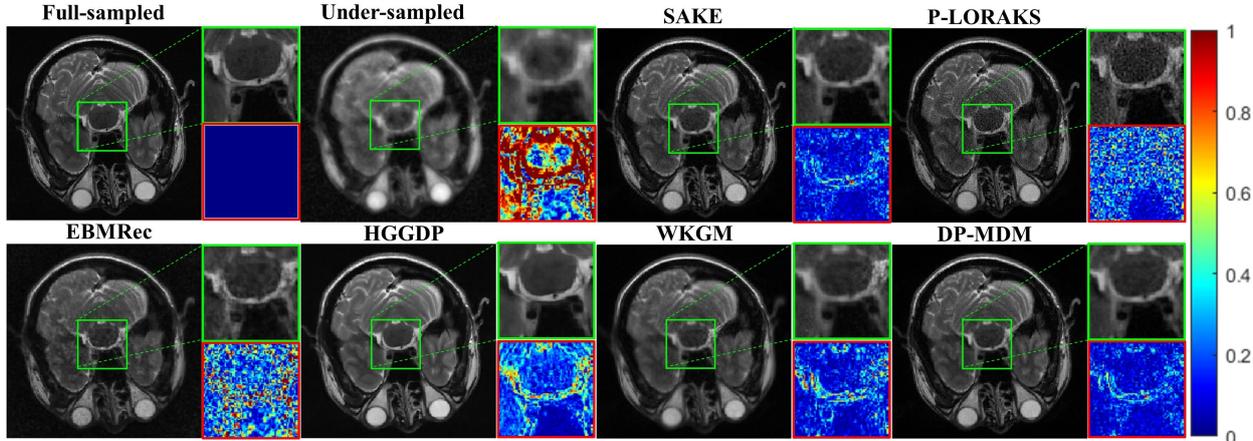

**Fig. 7.** Complex-valued reconstruction results at $R$=12 using 2D Poisson sampling with 12 coils. Top from left to right: Full-sampled, Under-sampled, reconstruction by SAKE and P-LORAKS; Bottom from left to right: reconstruction by EBMRec, HGGDP, WKGM, and DP-MDM (OURS). Green and red boxes illustrate the zoom-in results and error maps, respectively.

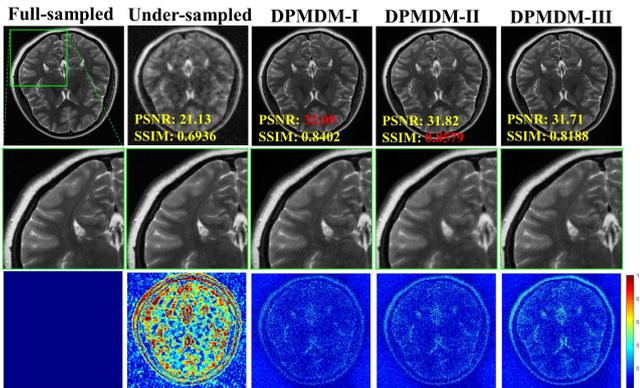

**Fig. 8.** Reconstruction results using various virtual binary masks in Fig. 5 at $R$=6 of the 2D Poisson sampling.

*Performance of DP-MDM Using Various Virtual Binary Masks:* To further compare the reconstruction results of the proposed DP-MDM using three virtual binary mask shapes in Fig. 5, the reconstructed images are visualized in Fig. 8. Each shape emphasizes different information relationships. Compared to the circular virtual mask of DPMDM-I, the radial virtual mask of DPMDM-II removes more high-frequency information in all directions, representing an intersected relationship between the effective information $E_1$ and $E_2$. Besides, the random masks of DPMDM-III removes frequency information through the rectangular shape, representing a disjoint relationship between the effective information $E_1$ and $E_2$. The results show that the performance of circular and radial masks, which involve contained and intersected relationships, respectively, outperform the disjoint random mask of DPMDM-III. Specifically, DPMDM-I re-veals the best reconstruction results in terms of PSNR, while DPMDM-II performs best in terms of SSIM. This implies that employing circular and radial virtual masks leads to effective reconstructed images. Hence, the effective information $E_1$ and $E_2$ intersect are more beneficial for reconstruction compared to disjoint relationships. Additionally, visually inspecting error maps reveals that the reconstruction results of DPMDM-I surpass those of the other two models. Thus, the design of the circular virtual mask shape is better suited for MRI image reconstruction in k-space domain.

*Comparison of Latest Masked Diffusion Models:* To comprehensively evaluate the efficacy of the proposed DP-MDM algorithm, a thorough comparison with state-of-the-art mask diffusion models is indispensable. A comprehensive evaluation is conducted against the latest CM-DM [38] method. Depicted in Fig. 9 are the visual and quantitative results of different models in reconstructing MRI images under a Poisson sampling rate of $R$=8. The data reveal that DP-MDM outperforms the CM-DM model by 1.18 dB in terms of PSNR and exceeds it by 4.92% in SSIM. Visually, observations indicate that CM-DM tends to acquire reconstructions with more noise, whereas DP-MDM successfully reconstructs the structural content of the image, particularly numerous fine details. This outcome is attributed to the utilization of circular virtual masks distinct from CM-DM, which are designed to better align with the distribution of k-space data. Additionally, employing a more intricate detail feature enhancement strategy than CM-DM enables DP-MDM to attain more optimal reconstruction results.

in both PSNR and SSIM, increasing from 29.59 dB/0.7507 to 35.18 dB/0.8990 (+5.59 dB/0.1483), respectively. Encouraged by these findings, we continue to add an extra diffusion model, resulting in further improvements from 35.18 dB/0.8990 to 38.05 dB/0.9094 (+2.87 dB/0.0104). Although the rate of improvement gradually slows down, the enhancements remain acceptable. However, upon comparing the performance of the 3-GMs and 4-GMs setups, the difference in metrics becomes marginal (+0.01 dB/0.0010), suggesting that the cost-effectiveness of adding another diffusion model diminished. These observations are consistent when applying an acceleration factor of $R$=10 as well.

TABLE III
PSNR AND SSIM MEASURES FOR DIFFERENT DIFFUSION NUMBERS WITH VARYING ACCELERATE FACTORS.

| Fast-MRI | 1-GM | 2-GMs | 3-GMs (DP-MDM) | 4-GMs |
|---|---|---|---|---|
| Poisson $R$=8 | 29.59/0.7507 | 35.18/0.8990 | 38.05/0.9094 | 38.06/0.9104 |
| Gap* | ------ | 5.59/0.1483 | 2.87/0.0104 | 0.01/0.0010 |
| Poisson $R$=10 | 29.16/0.7449 | 35.18/0.8995 | 37.62/0.9001 | 37.67/0.9033 |
| Gap* | ------ | 6.02/0.1546 | 2.44/0.0006 | 0.05/0.0032 |

*Gap represents the difference between the corresponding indicator values of the current column and the subsequent column.

Fig. 10 displays reconstructed images generated using varying numbers of diffusion models. With an increased number of diffusion models, the fading effect in the enlarged ROI of the residual map diminishes. The minimal difference between the 3-GMs and 4-GMs methods suggests that employing three diffusion models strikes a balance between information acquisition complexity and computational expenses. Insufficient diffusion models fail to yield ample information for high-quality image reconstruction, whereas an excess of models escalates data processing and computational overheads. Hence, utilizing the 3-GMs enables the acquisition of adequate information to unveil organizational and detail characteristics while avoiding redundant information and unnecessary complexity.

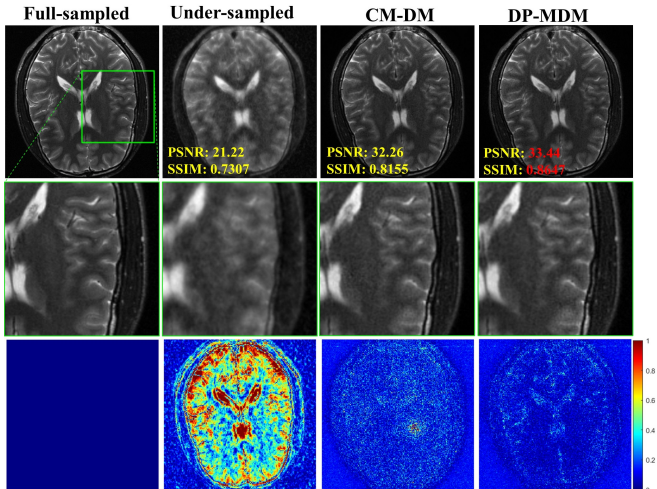

**Fig. 9.** Reconstruction results using CM-DM and DP-MDM at $R$=8 of the 2D Poisson sampling. The second row shows the enlarged view of the ROI (indicated by the green box in the first row), and the third row shows the error maps of the reconstruction.

### C. Ablation Study

To explore the optimal number of diffusion generative models to employ, ablation experiments are conducted wherein the number of GM is varied, including 1, 2, 3, and 4. The outcomes from the ablation study on 2D Poisson sampling with 20 coils are summarized in Table III. The results demonstrate that the top row indicates the number of diffusions employed, with "3-GMs" denoting the application of three diffusion models for MRI reconstruction. The third and fifth columns represent the difference in PSNR and SSIM between the current model and its predecessor with fewer diffusion models. Notably, upon adding one additional diffusion model to the 1-GM setup, there is an improvement

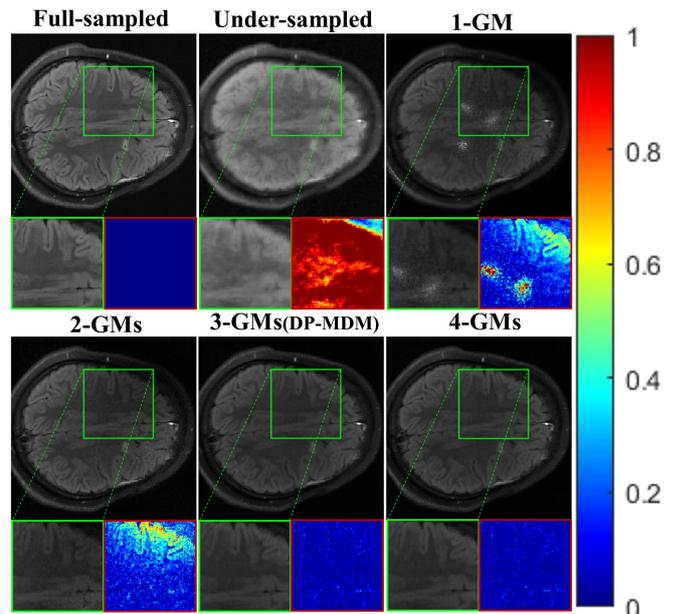

**Fig. 10.** Complex-valued reconstruction results at $R$=8 using 2D Poisson sampling with 20 coils. Green and red boxes illustrate the zoom-in results and error maps, respectively.

## V. Discussion

In this section, the main focus is on investigating the impact of parameter and model selection, with the aim of optimizing the MRI reconstruction process and improving the quality of the resulting reconstructions.

*Hyper-parameter Analysis:* We investigate the impact of the window size of virtual masks on the diffusion process of $GM_2$ and $GM_3$, as shown in Table IV. The study focuses on adjusting the window size of masks to extract high-frequency information at various levels, thereby enhancing the preservation of details during MRI reconstruction. Experiments are designed with different diameters for the window size of masks, such as "M10,20" representing the windows with 10 and 20 diameters of the circular masks used in $GM_2$ and $GM_3$, respectively. The results reveal that setting the window size to "M20,50" led to the evaluation metrics PSNR and SSIM reaching their maximum values of 34.76 dB and 0.8908, indicating superior reconstruction quality and more effective preservation of details at this specific point. DP-MDM works best with moderate sizes. However, we also observe that regardless of the window size, the changes in PSNR and SSIM remained marginal, with a difference of approximately 0.3 dB. This finding suggests that the proposed model demonstrates good robustness for its parameters.

TABLE IV
COMPARISON OF PSNR AND SSIM OF DIFFERENT WINDOW SIZES IN POISSON SAMPLING PATTERNS.

| Poisson R=6 | | Test 5 | Test 12 |
|---|---|---|---|
| **Window Sizes** | *M10,20\** | 34.45/0.8894 | 32.82/0.8807 |
| | *M20,50* | **34.76/0.8908** | **33.12/0.8833** |
| | *M50,70* | 34.72/0.8908 | 33.08/0.8828 |
| | *M70,120* | 34.58/0.8891 | 32.91/0.8809 |

*M10,20 represents windows with diameters of 10 and 20 for circular masks used in GM2 and GM3, reflecting the sizes of $m_1$ and $m_2$ relative to the parameter $a$ in Eq. (8).

*Connected Patterns:* To explore the impact of various combinations of multiple diffusion models on image quality, four combination patterns are analyzed as depicted in Fig. 3, and presented the findings in Table V. From the results, we observe that the "(a) Cascade" method accomplished the best reconstruction result, while the "(c) Hybrid 2" method produce the poorest reconstruction outcome, displaying a PSNR difference of approximately 3.00 dB between them. Notably, the reconstruction results of the "(a) Cascade" and "(b) Hybrid 1" methods are remarkably similar, with a PSNR difference of less than 0.1 dB. This signifies the crucial role of the $GM_1$ model, which incorporates prior information regarding the distribution of global structural data. Meanwhile, the $GM_2$ and $GM_3$ models, integrating local data distribution information, play a crucial role in preserving high-frequency detailed features.

TABLE V
COMPARISON OF PSNR AND SSIM OF FOUR DIFFERENT COMBINATION PATTERNS IN FIG. 3.

| 2D Poisson R=6 | | Test 5 | Test 12 |
|---|---|---|---|
| **Patterns** | (a) *Cascade* | **34.76/0.8908** | **33.12/0.8833** |
| | (b) *Hybrid 1* | 34.75/0.8908 | 33.04/0.8830 |
| | (c) *Hybrid 2* | 31.72/0.8700 | 31.91/0.8722 |
| | (d) *Parallel* | 32.22/0.8537 | 31.23/0.8394 |

Overall, the "(a) Cascade" connection plays a dominant role in the reconstruction stage, facilitating close interconnection and progressive layer-by-layer integration of multiple diffusion models. Finally, the "(a) Cascade" combination pattern is selected as the optimal choice for this work.

## VI. Conclusion

In this work, we presented a novel comprehensive detail-preserving reconstruction method using multiple diffusion models. Specifically, multiple diffusion models were utilized to extract structural and detailed feature information from k-space. At the same time, binary modal masks were employed to refine the value range of the k-space data with highly adaptive center windows, which allows the model to focus its attention more efficiently and maximize the reconstruction performance of the model. Additionally, the efficiency of learning data distribution in the diffusion model was enhanced through the implementation of an inverted pyramid structure. The structure systematically reduced image information from the top to the bottom, resulting in a cascade and sparse representation. Finally, we validated the proposed method on both clinical and public datasets and demonstrated its superiority over other methods in reconstructing feature details. However, the large number of iterations in the diffusion model led to prolonged reconstruction time. In the future, we aim to enhance not only the accuracy of image reconstruction but also the generation speed of the diffusion model.